\documentclass[conference]{IEEEtran}
\IEEEoverridecommandlockouts
\usepackage{cite}
\usepackage{amsmath,amssymb,amsfonts}
\usepackage{algorithmic}
\usepackage{graphicx}
\usepackage{textcomp}
\usepackage{xcolor}
\def\BibTeX{{\rm B\kern-.05em{\sc i\kern-.025em b}\kern-.08em
    T\kern-.1667em\lower.7ex\hbox{E}\kern-.125emX}}

\usepackage{times}
\usepackage{epsfig}
\usepackage{graphicx}
\usepackage{amsmath}
\usepackage{amssymb}
\usepackage{graphicx}
\usepackage{lscape}
\usepackage{comment}
\usepackage{caption}
\usepackage{subcaption}
\usepackage{float}
\usepackage{multirow}
\usepackage[utf8x]{inputenc}
\usepackage[breaklinks=true,bookmarks=false]{hyperref}

\begin{document}

\title{Ultra Sharp : Study of Single Image Super Resolution using Residual Dense Network}
 \author{Karthick Prasad Gunasekaran \\
 University of Massachusetts\\
Amherst, USA\\
 {\tt\small kgunasekaran@umass.edu}
 }


\maketitle
\begin{abstract}
For years, Single Image Super Resolution (SISR) has been an interesting and ill-posed problem in computer vision. The traditional super-resolution (SR) imaging approaches involve interpolation, reconstruction, and learning-based methods. Interpolation methods are fast and uncomplicated to compute, but they are not so accurate and reliable. Reconstruction-based methods are better compared with interpolation methods, but they are time-consuming and the quality degrades as the scaling increases. Even though learning-based methods like Markov random chains are far better than all the previous ones, they are unable to match the performance of deep learning models for SISR. This study examines the Residual Dense Networks architecture proposed by Yhang et al. \cite{srrdn}   and analyzes the importance of its components. By leveraging hierarchical features from original low-resolution (LR) images, this architecture achieves superior performance, with a network structure comprising four main blocks, including the residual dense block (RDB) as the core. Through investigations of each block and analyses using various loss metrics, the study evaluates the effectiveness of the architecture and compares it to other state-of-the-art models that differ in both architecture and components.

\end{abstract}

\section{Introduction}
 Single Image Super Resolution (SISR) is the process by which high-resolution (HR) images are recovered from low-resolution (LR) images. It is one of the most important problems in computer vision and has a wide range of applications, from security and surveillance purposes to medical imaging scenarios. The problem can be dealt with in a variety of ways, from classical computer vision techniques to using deep learning to solve it. The classical super resolution techniques include interpolation based methods \cite{interpolation} such as nearest neighbors, bilinear, bicubic interpolation, and reconstruction-based methods. However, those methods are unable to perform as well as the learning-based methods. Given the increasing amount of data available, learning-based methods are able to achieve better results, and the prediction time is comparatively shorter. Various deep learning based methods have been proposed for solving the image superresolution problem.
 
The first basic convolutional neural network-based method was proposed by Dong et al. \cite{srcnn} in 2014 which performed much better than traditional methods. From then on, various different approaches and architectures were proposed, differing in terms of network architecture, loss functions, and learning principles. Kim et al.\cite{kim2015accurate} in the following year, proposed a very deep convolution network inspired by VGG-net. This way, they were able to capture the contextual information over large regions of the images. However, the models proposed should have different architectures for different scales. Inspired by ResNet architecture, Lim et al. \cite{lim2017enhanced} proposed a residual learning-based approach where high-resolution images are produced from different scaling factors by using a single model. Tai et al. \cite{tai2017memnet} proposed MemNet, which tackles the long-term dependency problem in deep models. MemNet uses memory blocks to explicitly mine persistent memory through an adaptive learning process. Memory blocks help in adaptively controlling how much should be remembered from previous states. Tong et al. \cite{tongtong} proposed a dense skip connections-based approach where the feature maps of each layer are added to the feature maps of every other layer by effectively combining low-level features with high-level features and eliminating the vanishing gradient problem. Deconvolution layers are added to the network to learn the upsampling filters and speed up reconstruction.  

 \begin{figure*}[ht]
   \includegraphics[width=\textwidth]{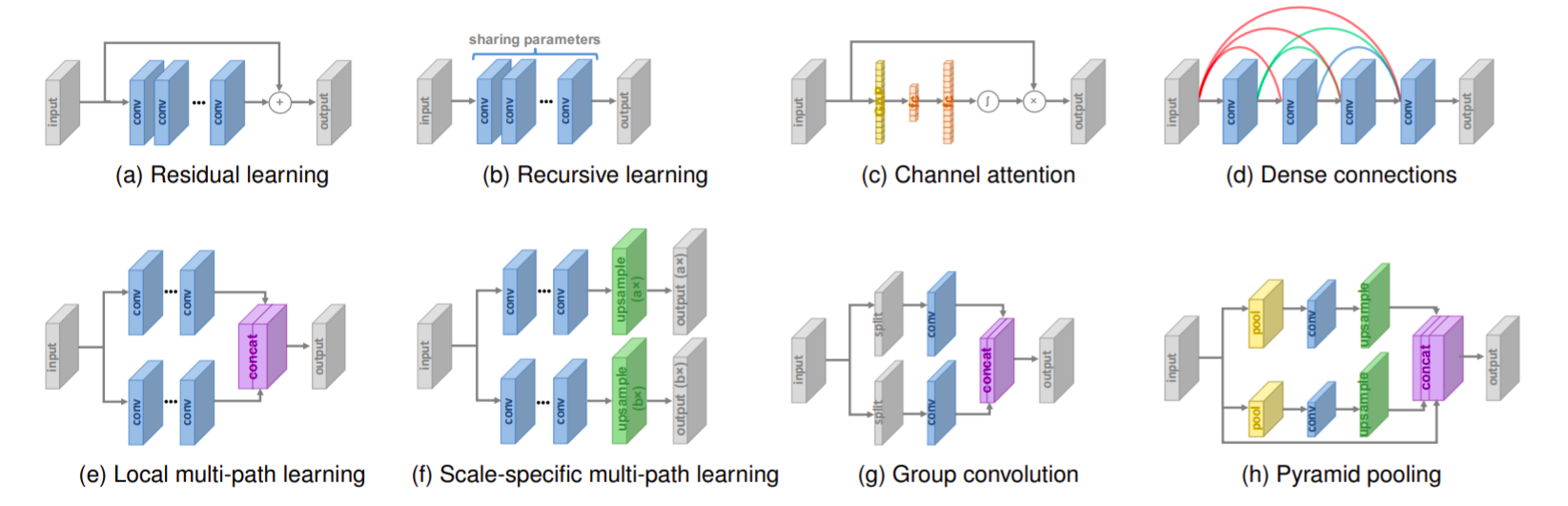}
    \caption{Various Design strategies \cite{sr_survey}}
    \label{fig:litmain1}
\end{figure*}

However, there are problems with these approaches. These methods fail to fully make use of the hierarchical features for reconstruction. Also, the methods don't extract multi level features from the images, and in some methods, upsampling is performed as pre-processing from low resolution, which greatly misses much of the information and increases the time complexity as well. To overcome these problems, a new residual dense network was proposed by Zhang et al \cite{srrdn}. This network fully makes use of hierarchical features, with residual dense blocks (RDB) forming the core of the architecture. The network structure will consist of four main blocks: shallow feature extraction network, residual dense block, dense feature fusion block, and up sampling block. The shallow feature extraction module will extract shallow features by using convolutional layers. Residual block has densely connected layers, local feature fusion (LFF), and uses continuous memory (CM) mechanisms. Dense feature fusion blocks use global feature fusion and global residual learning. The global feature fusion module adaptively fuses hierarchical features from all RDBs in low-resolution space. The layers in RDB consist of densely connected layers and local feature fusion. The contiguous memory mechanism is also implemented in the RDB.

In this work, We evaluate and study the following,
 \begin{itemize}
 
 \item Evaluate the new architecture which makes use of hierarchical features called Residual Dense Network is presented for single image super resolution.

 \item Studying the entire novel residual dense architecture framework  \cite{srrdn}.
 \item Finding the relative efficiency/importance of the blocks proposed by Residual Dense Network \cite{srrdn}.
 \item Make a comparison with other state of the art models with same no of epochs however having different hyper parameters. 
 \end{itemize}

\section{Background/Related Work}
  Super resolution is an ill-posed problem, there are many different problems, such as designing up-sampling blocks, deciding the architecture to be used, etc. In this section, we will analyze the most prominent up-sampling strategies and architectures used by researchers over time to solve this problem.

\subsection{Upsampling Architectural Techniques}
In Super Resolution (SR), four different upsampling strategies are used in general to scale the low resolution image to map the high resolution image. The four model frameworks include pre-upsampling, post-upsampling, progressive upsampling, and iterative up-down upsampling.

\subsubsection{Pre-upsampling Architecture}
In this architecture, the upscaling is initially carried out using classical upsampling techniques such as bicubic interpolation. This helped the model reduce complexity since the CNN could just refine the up-sampled image to obtain the HR image. The complex upsampling task was carried out using classical techniques. This setup was initially adopted by \cite{srcnn} for SR-CNN and was widely used earlier during the initial development of SR architectures using the CNNs. Figure \ref{fig:upsample_archi}(a) shows  pre-upsampling architecture. However, this approach has been replaced by learning based approaches in recent times. The disadvantage of this is that there is no learning taking place when the LR images are upscaled to HR images.
 \subsubsection{Post-upsampling Architecture}
In post-upsampling architecture, multiple learnable layers are added at the end of the SR architecture where upsampling is performed. This kind of framework was initially proposed by Dong et al. \cite{dong2016accelerating} where he performs most of the mappings in LR space.
Figure \ref{fig:upsample_archi}(b) shows a sample post upsampling architecture. Since most of the features are extracted in LR space, the complexity is low and the computational cost is also low. This increases the performance of the network and is widely used in current SR research. 

\subsubsection{Progressive upsampling Architecture}
 Progressive upsampling was suggested to overcome the disadvantages of pre- and post-upsampling. In pre-upsampling, the upscaling is done in a single step, which inhibits learning, while in post-upscaling, a new network is needed that cannot perform multiscale super resolution. Figure \ref{fig:upsample_archi}(c) shows an example of progressive upsampling architecture. Lai et al. proposed a Laplacian pyramid network \cite{lai2017deep} to overcome these difficulties. This model progressively reconstructs high resolution images. This model, however, has its own problems, such as complicated architecture and design. They also required advanced training strategies and guidance.

 \subsubsection{Iterative upsampling Architecture}
\begin{figure}[ht]
 \centering
     \begin{subfigure}[b]{0.46\textwidth}
           \centering
         \includegraphics[width=\textwidth]{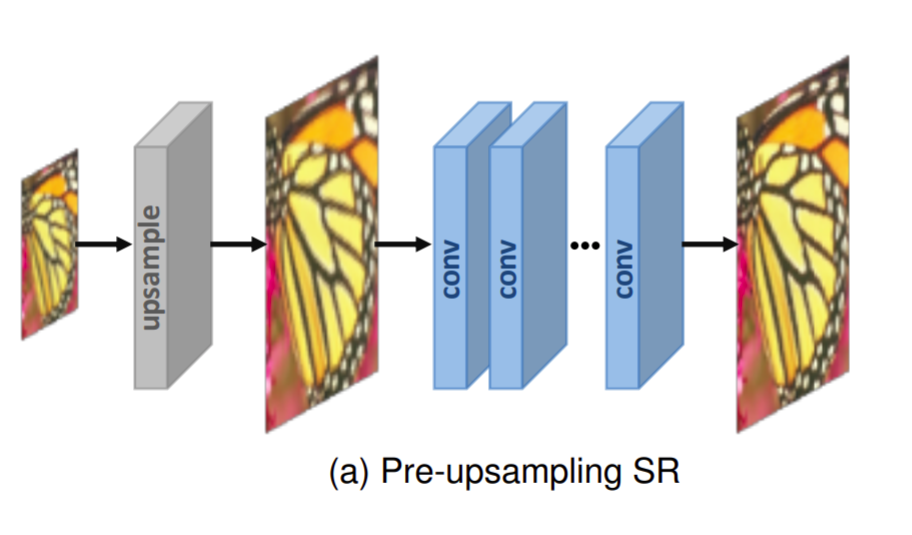}

         \label{fig:img_preup}
     \end{subfigure}
     \centering
     \begin{subfigure}[b]{0.46\textwidth}
           \centering
         \includegraphics[width=\textwidth]{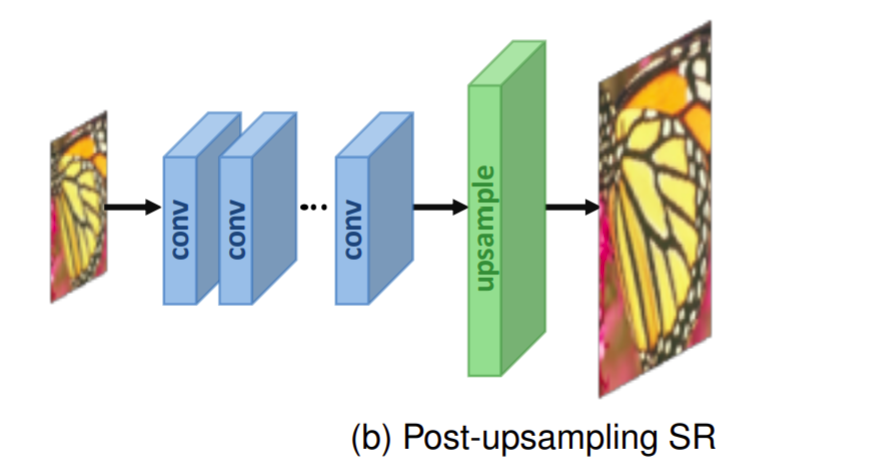}
     
         \label{fig:img_post_up}
     \end{subfigure}
     \centering
      \begin{subfigure}[b]{0.46\textwidth}
           \centering
         \includegraphics[width=\textwidth]{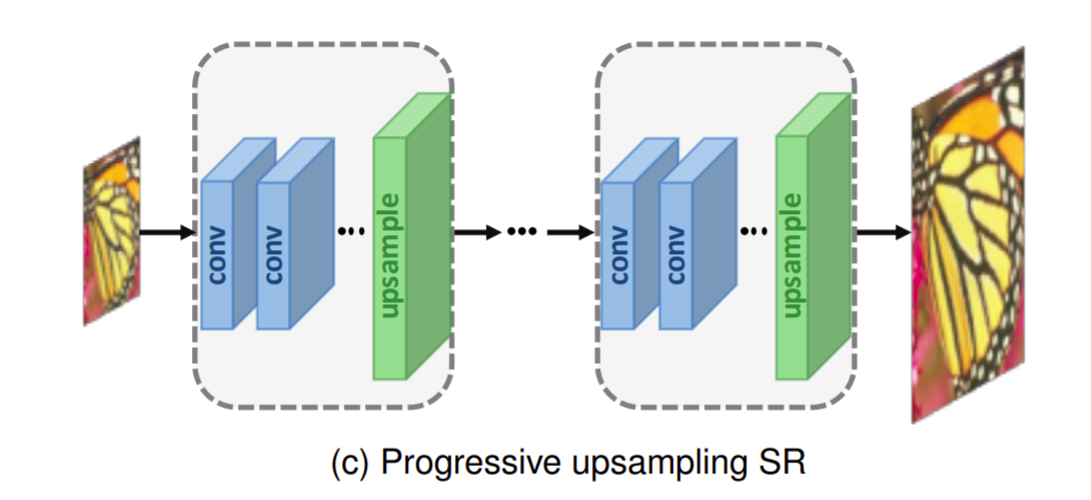}
      
         \label{fig:img_pro_up}
     \end{subfigure}
     
         \centering
      \begin{subfigure}[b]{0.46\textwidth}
           \centering
         \includegraphics[width=\textwidth]{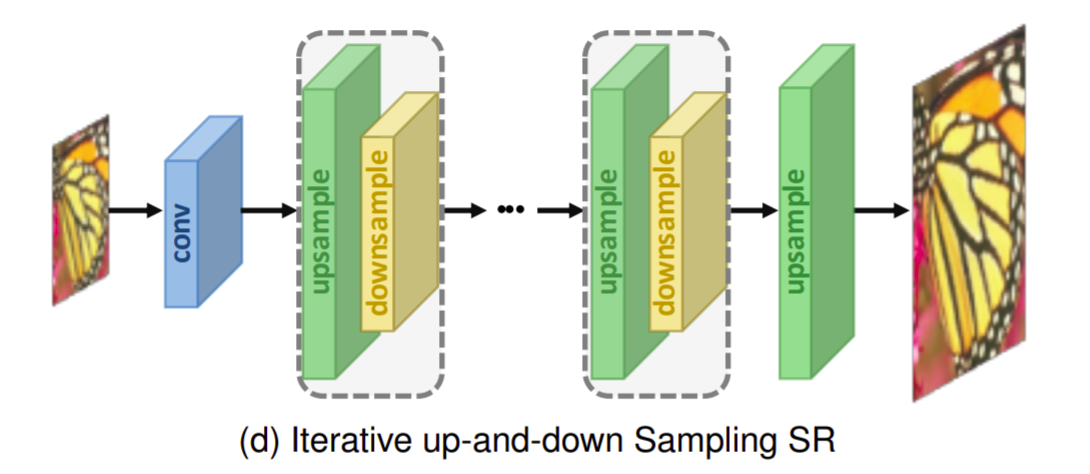}
      
         \label{fig:img_iterative_up}
     \end{subfigure}
     
       \caption{Different Up-sampling architectures 
       \cite{sr_survey}}
        \label{fig:upsample_archi}
\end{figure}

In iterative upsampling architecture,the mutual dependency of LR and HR images is used. This is considered a reconstruction problem where the LR image is iteratively converted into an HR image and compared with the HR image to compute the loss. However, this approach brings complication to the design since different training and testing frameworks have to be present. This is the most recently proposed architecture that has been under study in recent years. Figure \ref{fig:upsample_archi}(d) \subref{fig:img_iterative_up} shows a sample iterative upsampling architecture.
    
\subsection{Design}
In SR architectures, various design strategies are being carried out to improve efficiency. Commonly used network designs are explored and discussed in this section.

\subsubsection{Residual Learning}
Inspired by ResNet \cite{resnet} architecture, residual learning is applied at different levels. Figure \ref{fig:litmain1}(a) shows a sample residual design. There are two main methods of residual learning: global residual learning and local residual learning. These are employed by making a direct connection from the input block to the output block. 

\subsubsection{Recursive Learning}
In recursive learning, the same module is applied multiple times in a recursive manner, which helps in achieving a wider and larger receptive field. This also helps in extracting higher level features helps in achieving a wider and larger receptive field. This also helps in extracting higher-level features. However, there are problems like vanishing and exploding gradients in this. A sample of  which is shown in \ref{fig:litmain1}(b). 

\subsubsection{Multi-Path Learning}
As shown in  \ref{fig:litmain1}(e) and (f) multi-path learning involves learning features in multiple paths where, in each path, different operations are performed. For SR purposes, they are divided into global, local, and scale-specific multi-path learning. In scale specific learning, there are multiple different paths for different scales of images.
\begin{figure*}[ht]
\includegraphics[width=\textwidth]{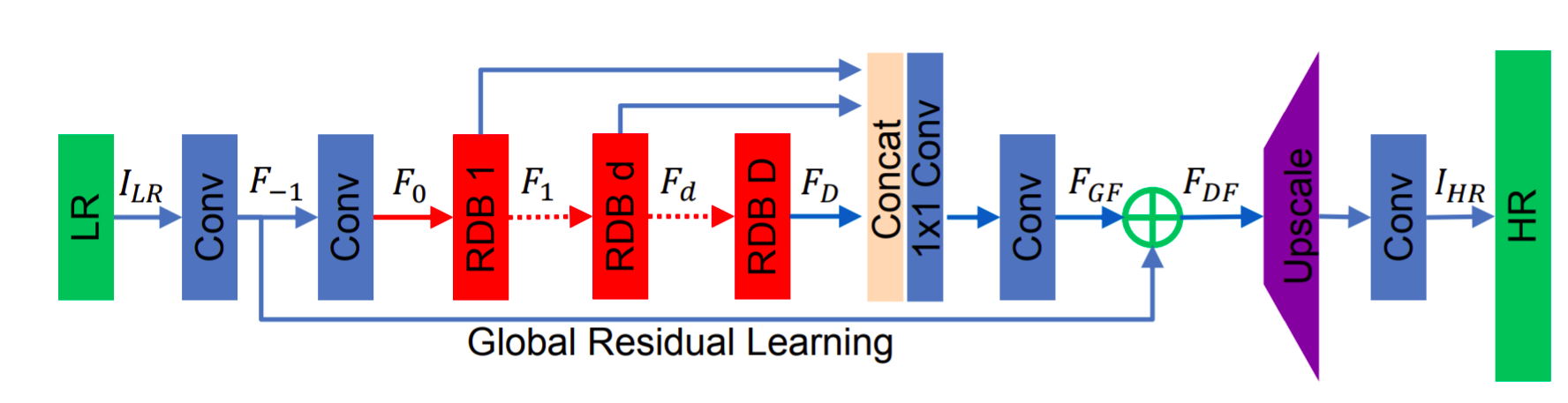}
\caption{Architecture of Residual Dense network  \cite{srrdn}}
\label{fig:architecture_rdn}
\end{figure*}
\subsubsection{Dense layer connection}
The dense layer connections were inspired by the DenseNet proposed by Huang et al. \cite{densenet}. In each layer, connections are made with all the other previous layers by having the feature maps of the preceding layers input here. This connection helps with better signal propagation and feature reuse. Figure \ref{fig:litmain1}(d) shows dense connections.

\subsubsection{Channel Attention}
In the channel attention mechanism, the inputs are fused into a channel descriptor, where global average pooling is performed. Then the descriptors are directly used by two fully connected layers to produce scaling factors for each channel. The channel attention mechanism is shown in Figure  \ref{fig:litmain1}(c).
\section{Approach}

\begin{figure*}[ht]
   \includegraphics[width=\textwidth]{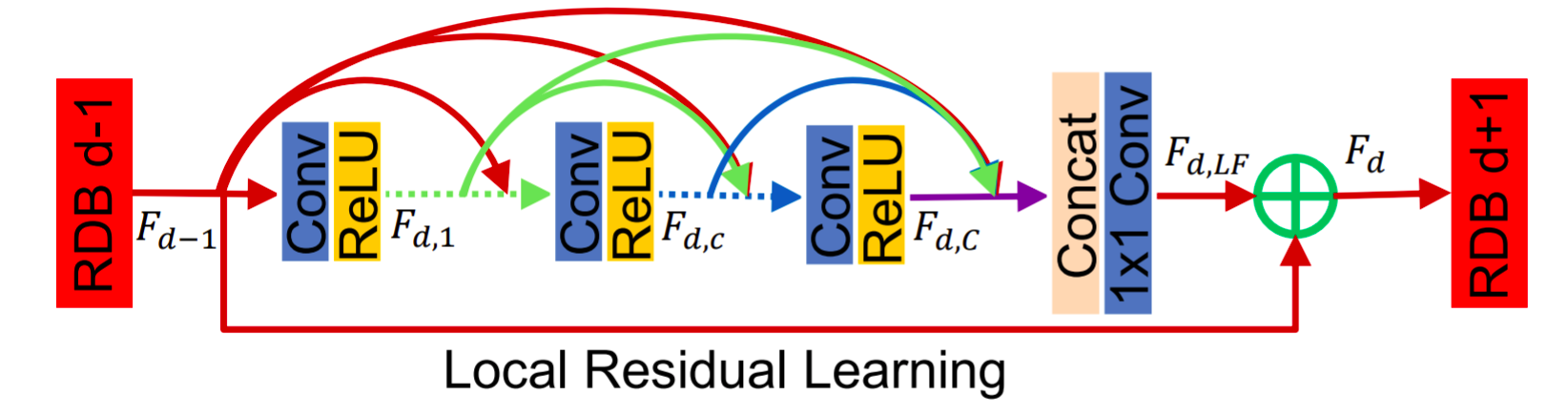}
    \caption{Architecture of Residual dense block \cite{srrdn}}
    \label{fig:architecture_rdb}
\end{figure*}

\subsection{Network Architectures:}
  The main residual dense network architecture consists of four main components as shown in the Figure \ref{fig:architecture_rdn}. 
  They main blocks are 
  \begin{itemize}
      \item Shallow Feature Extraction block
       \item Residual Dense Block 
      \item Dense Feature Fusion block 
      \item Upsampling block
  \end{itemize}

\subsubsection{Shallow Feature Extraction(SFE) block}
The shallow feature extraction block consists of convolutional layers where the features are extracted in low resolution (LR) space. The features extracted are fed into each of the residual dense block layers and finally used in residual learning for global feature fusion. This layer is able to achieve a good high level representation of the images. It consists of two convolutional layers, where the output of the first layer is fed into the second convolutional layer.
 
 Let $Img_{LR}$ be the input low resolution image and $Conv$ be the convolutional operation in feature extraction layer. 
 \begin{equation}
  F_{-1}= Conv_1(Img_{LR})
 \end{equation}
 where $F_{-1}$ are the features extracted from input low resolution image 
  \begin{equation}
  F_{0}= Conv_2(F_{-1})
 \end{equation}

Here, $F_{0}$ be the features extracted from this block which will be fed into residual dense block.

\subsubsection{Residual Dense Block}
 The Residual Dense Block(RDB) performs the Local residual learning and Local feature fusion  which helps in extracting the local dense features and providing better context locally.
 
Multiple residual dense blocks are implemented in a feed-forward manner, taking the output of the previous block as the input to the current block. Each block consists of three feed-forward convolutional layers, followed by a non-linear layer, and  finally all the layers within a block are fused and residual input is applied. Figure \ref{fig:architecture_rdb} clearly shows the architecture of the residual dense block (RDB). Each residual block makes use of the output from the previous block, as shown in the below equation. 

Let $F_{d-1}$ be the features from the previous layer.
  \begin{equation}
     F_{d} =  Conv_{RDB,1}(F_{d-1})
 \end{equation}

 Further residual dense blocks are processed in a feed forward manner as follows using the output from previous blocks,
  
   \begin{equation}
     F_{d} =  Conv_{RDB,d}(F_{d-1})
 \end{equation}
 
 \begin{equation}
    F_{d}    =Conv_{RDB,d}(Conv_{RDB,d-1}(...))
 \end{equation}
 
Within each residual dense block, local feature fusion is performed, where all the outputs of the previous layers are fused or concatenated adaptively together as shown in the figure \ref{fig:architecture_rdb} to be fed as input to the next layer within RDB. This means the output of all preceding layers of the current RDB and the output of the previous RDB block are given as input to the current layer within the RDB. A continuous memory mechanism is achieved since all the preceding RDB output is fed into each layer with the current RDB. This helps preserve the feed-forward tendency and extract dense features. The output and input for each layer in the RDB are given in the equation,
    \begin{equation}
   F_{d,l} =ReLU(Conv([F_{d-1},F_{d,1}...,F_{d,l-1}]))
   \end{equation}
   
  where $F_{d,l}$ and $F_{d-1}$,...,$F_{d,l-1}$ are the outputs and inputs of the current residual dense block and ReLU\cite{relu} is the non-linearity after the convolutional layer.
  
   Finally the concatenation of the outputs of RDB layers is carried out and 1*1 Convolutions are performed such that the feature maps size is reduced.  
 \begin{equation}
   F_{d} =Conv_{1*1}([F_{d-1},F_{d,1}...,F_{d,c}])
   \end{equation}
   
   After this the local residual learning is performed in which the the input to the RDB block is added to the output of the current RDB block. This helps in maintaining the information flow through out the network.
   
    \begin{equation}
   F_{d} = F_{d} + F_{d-1}
   \end{equation}
   
   Local residual Learning helps the network learn and remember local structure better resulting in better performance.

\subsubsection{Dense Feature Fusion(DFF) block}
In DFF, global feature fusion and global residual learning is performed. This helps in achieving the hierarchical features from the LR images in HR space. In this block all the outputs from RDB layers are concatenated and residual learning is performed from LR image input. 

  Global feature fusion is achieved by fusing or concatenating all the outputs from the RDBs and performing 1*1 convolution on top of it. A 3*3 convolutional filter is again applied on the output from 1*1 filter.
  \begin{equation}
   F_{GF} = Conv_{3*3}(Conv_{1*1}((F_1,....,F_D)))
   \end{equation}
  
   Global residual learning is performed in which the output from GFF is added up with the high level features of the LR input image. The output of which is fed into the upscaling layer.
     \begin{equation}
     F_{DFF} = F_{-1} + F_{GFF}
     \end{equation}
\subsubsection{Upscaling block}
 Upscaling to higher resolution space is carried out by using sub-pixel based convolutional neural network. The feature maps from the DFF which were extracted in low resolution space is  upscaled into HR space using an efficient sub pixel based approach as proposed by Shi et al. \cite{subpixelupscale}.

\section{Implementation} 
The code for the implementation can be found at the following github link:  \url{https://github.com/karthickpgunasekaran/SuperResolutionWithRDN}

\subsection{Dataset:}
Four different datasets were used for the overall training and testing of the network. For training, the DIV2K dataset was used. The DIV2K dataset consists of 800 training and 100 validation 2K resolution images. Degradation was carried out by using bicubic downsampling on the HR dataset to be used as inputs. The testing was carried out using the SET-5, SET-14, and URBAN-100 datasets. LR datasets were created for different scales, such as 2x, 3x, and 4x. 
   \newline
   
\begin{figure}[ht]
 \centering
     \begin{subfigure}[b]{0.14\textwidth}
           \centering
         \includegraphics[width=\textwidth]{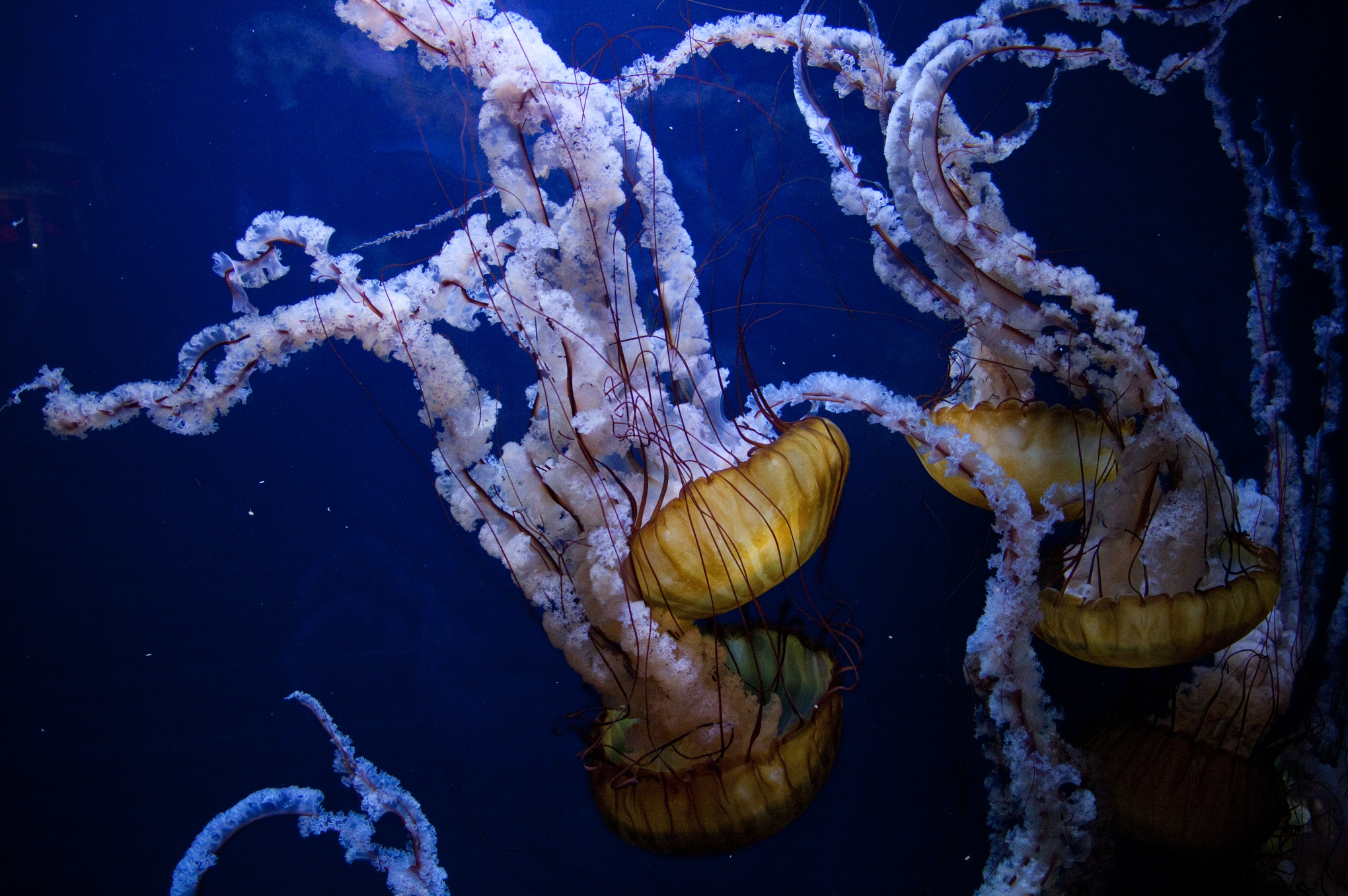}

         \label{fig:datasetapproachprop131}
     \end{subfigure}
     \centering
     \begin{subfigure}[b]{0.14\textwidth}
           \centering
         \includegraphics[width=\textwidth]{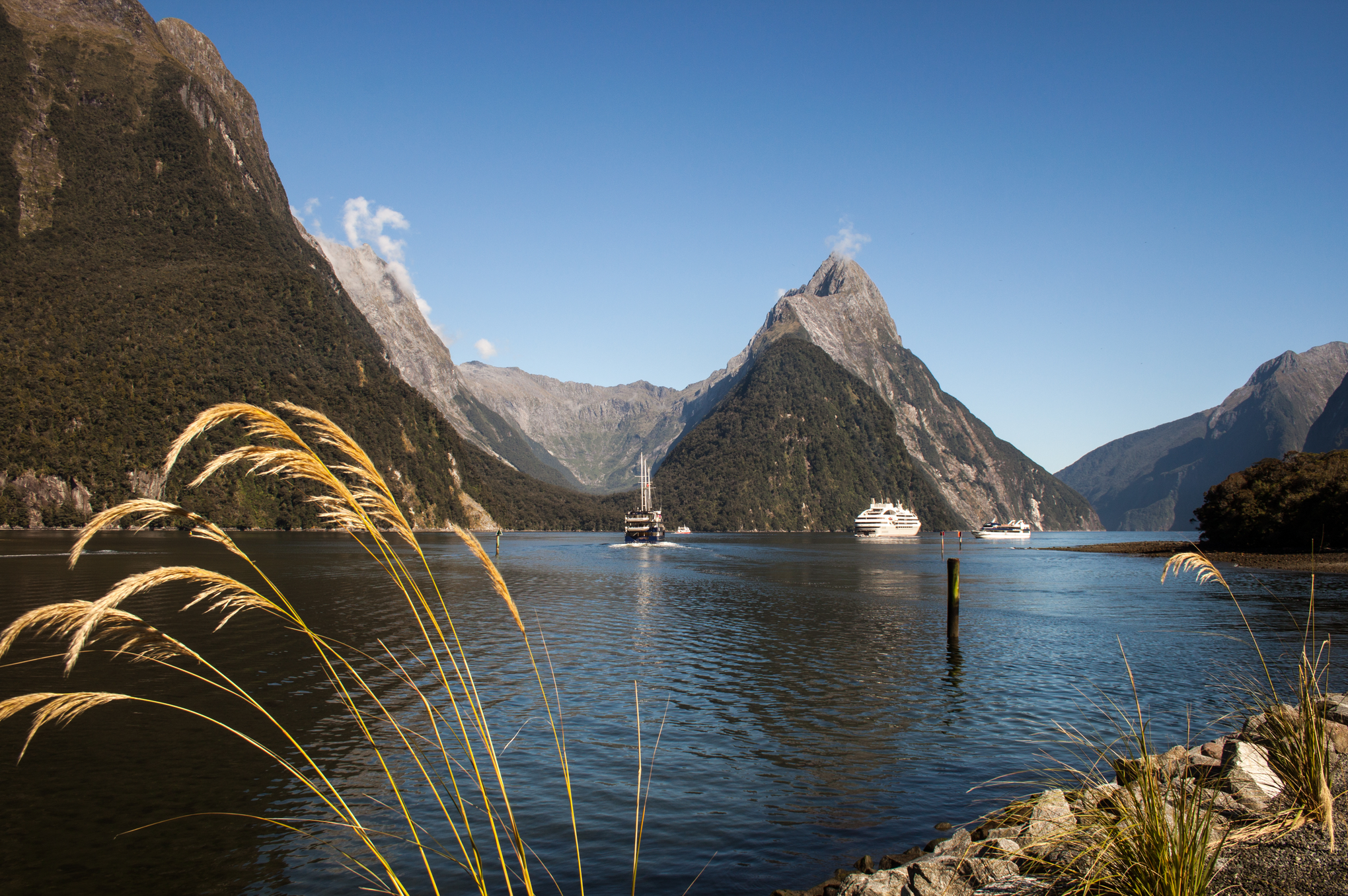}
     
         \label{fig:datasetapproachprop132}
     \end{subfigure}
     \centering
      \begin{subfigure}[b]{0.14\textwidth}
           \centering
         \includegraphics[width=\textwidth]{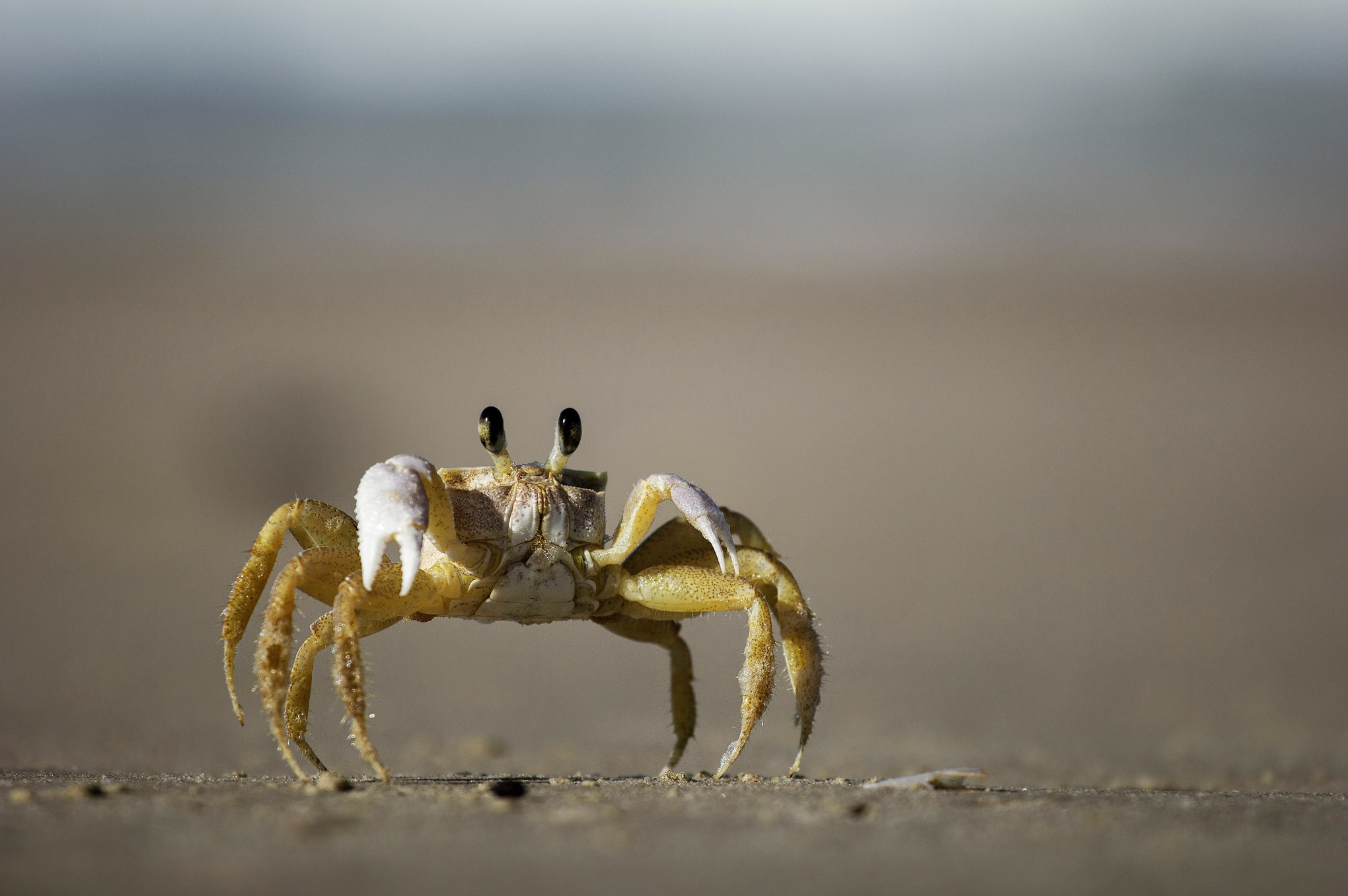}
      
         \label{fig:datasetapproachprop133}
     \end{subfigure}

     \centering
     \begin{subfigure}[b]{0.14\textwidth}
           \centering
         \includegraphics[width=\textwidth]{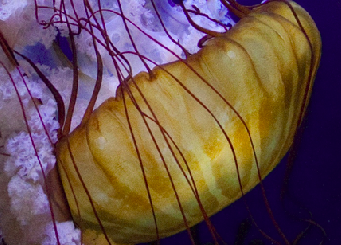}

         \label{fig:datasetapproachprop131}
     \end{subfigure}
     \centering
     \begin{subfigure}[b]{0.14\textwidth}
           \centering
         \includegraphics[width=\textwidth]{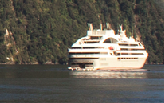}
     
         \label{fig:datasetapproachprop132}
     \end{subfigure}
     \centering
      \begin{subfigure}[b]{0.14\textwidth}
           \centering
         \includegraphics[width=\textwidth]{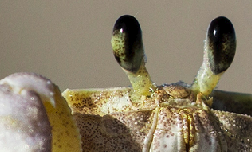}
      
         \label{fig:datasetapproachprop133}
     \end{subfigure}
     
          \centering
     \begin{subfigure}[b]{0.14\textwidth}
           \centering
         \includegraphics[width=\textwidth]{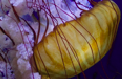}

         \label{fig:datasetapproachprop131}
     \end{subfigure}
     \centering
     \begin{subfigure}[b]{0.14\textwidth}
           \centering
         \includegraphics[width=\textwidth]{Images/Dataset/39_1x.png}
    
         \label{fig:datasetapproachprop132}
     \end{subfigure}
     \centering
      \begin{subfigure}[b]{0.14\textwidth}
           \centering
         \includegraphics[width=\textwidth]{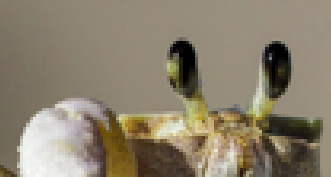}

         \label{fig:datasetapproachprop133}
     \end{subfigure}
            \centering
     \begin{subfigure}[b]{0.14\textwidth}
           \centering
         \includegraphics[width=\textwidth]{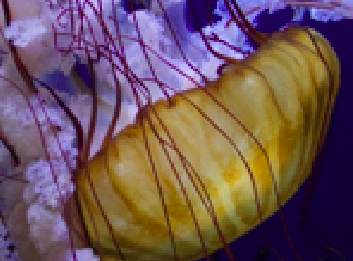}

         \label{fig:datasetapproachprop131}
     \end{subfigure}
     \centering
     \begin{subfigure}[b]{0.14\textwidth}
           \centering
         \includegraphics[width=\textwidth]{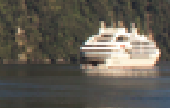}
    
         \label{fig:datasetapproachprop132}
     \end{subfigure}
     \centering
      \begin{subfigure}[b]{0.14\textwidth}
           \centering
         \includegraphics[width=\textwidth]{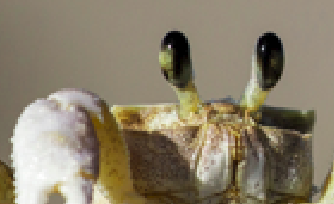}

         \label{fig:datasetapproachprop133}
     \end{subfigure}
                 \centering
     \begin{subfigure}[b]{0.14\textwidth}
           \centering
         \includegraphics[width=\textwidth]{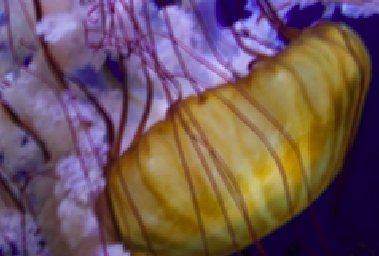}

         \label{fig:datasetapproachprop131}
     \end{subfigure}
     \centering
     \begin{subfigure}[b]{0.14\textwidth}
           \centering
         \includegraphics[width=\textwidth]{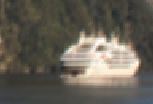}
    
         \label{fig:datasetapproachprop132}
     \end{subfigure}
     \centering
      \begin{subfigure}[b]{0.14\textwidth}
           \centering
         \includegraphics[width=\textwidth]{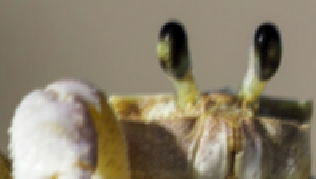}

         \label{fig:datasetapproachprop133}
     \end{subfigure}
       \caption{Various image scales  Row 1 - Original image, Row 2 - Original zoomed, Row 3 - 2x bicubic downsampling, Row 4 - 3x bicubic downsampling, Row 5 - 4x bicubic downsampling }
        \label{fig:datasetfig}
\end{figure}

   Figure \ref{fig:datasetfig}  shows different scaled versions of the dataset. In the first row, the whole original image of the dataset is displayed. To show the difference in resolution of the images in the next rows, zoomed versions of the images are presented. The second row is the original image, which is just zoomed in. The third row contains images that are 2x downsampled by bicubic downsampling. The fourth and fifth rows show images that are 3x and 4x downsampled by bicubic downsampling.

\subsection{Loss Metrics:}
The network uses different loss metrics training and evaluation. The training was carriedd out using L1 loss metric. While
Structure Similarity Index Matrix (SSIM) \cite{ssim} and Peak Signal to Noise Ratio(PSNR) were used as the loss functions for the evaluation. SSIM gives a good measure of how similar two images are  in terms of brightness,contrast and structure while PSNR focuses on noise on the image. 
 \begin{figure}[H]
   \includegraphics[width=0.42\textwidth]{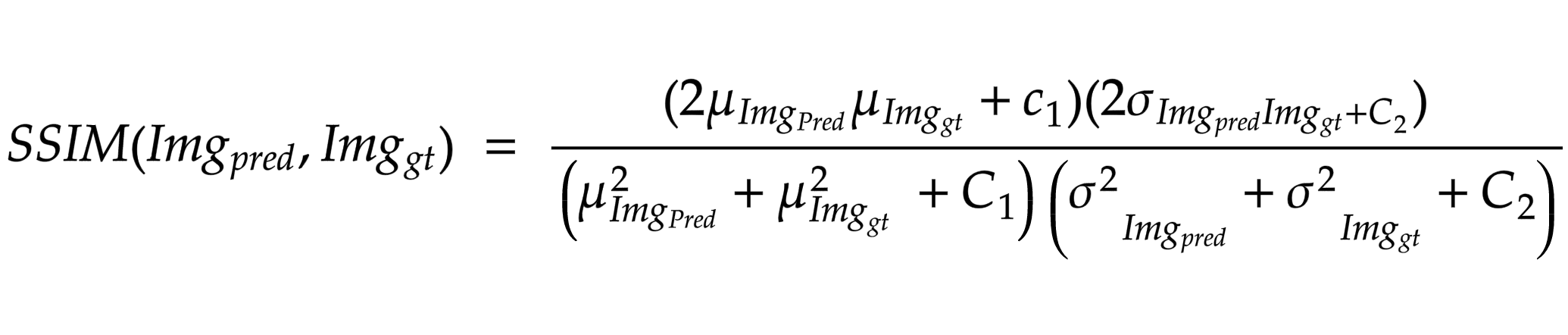}
   
    \label{fig:ssim}
\end{figure}
 \begin{figure}[H]
   \includegraphics[width=0.42\textwidth]{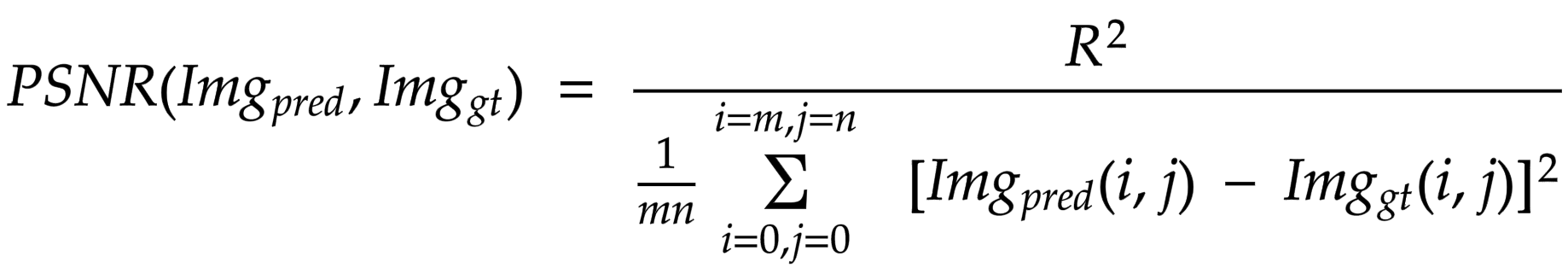}
   
    \label{fig:psnr}
\end{figure}
 \begin{figure}[H]
   \includegraphics[width=0.42\textwidth]{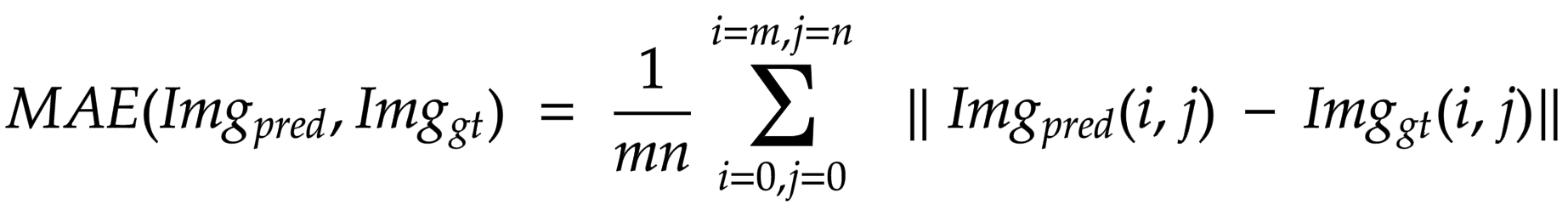}
   
    \label{fig:mae}
\end{figure}
 where $Img_{pred}$ and $Img_{gt}$ are the predicted and ground truth HR images and $R$ is the maximum fluctuation in the input image data.  
\subsection{Setting:}
The batch size of 8 HR and LR images where used for training the model and batch size of 4 was used for testing the model. The weights were initialized by kaiming he method \cite{kaiming}. Adam optimizer \cite{adam} was used to perform gradient updates. The regularization was set to 1e-8. The learning rate was initially set to 1e-4 and was reduced every 15 epochs by 2. The model was trained on Google lab for 200 iterations. The model took 14 1/2 hours to train. Since Colab supports only approximately 12 hours of continuous usage model was saved in between and continued. 
\subsection{Results:}
\begin{table}[ht] 
\centering

 \caption{\label{tab:table_scale}Comparison of model performance among different scales} 
 
 \begin{tabular}{|l | c| c| c| c|c| c| c|} 
 \hline
\multirow{2}{*}{Dataset} & \multicolumn{2}{c|}{2x}  & \multicolumn{2}{c|}{3x} & \multicolumn{2}{c|}{4x}  \\ 
 
  & PSNR & SSIM & PSNR & SSIM & PSNR & SSIM \\
 \hline

Set-5 & 30.67 &	0.91  &	28.05 &	0.88  &	26.19 &	0.85 \\
\hline
Set-14 & 28.56 &	0.87  &	27.78 &	0.83 &	24.21 &	0.79\\
\hline
Urban-100   & 28.93 &	0.88 & 26.3  &	0.83  &	24.02 &	0.77\\
\hline

\end{tabular}
\end{table}

\begin{table*}[ht] 
\centering

 \caption{\label{tab:table_model}Comparison of model performance with different models} 
 
 \begin{tabular}{|l | c| c| c| c|c| c| c| c| c|} 
 \hline
\multirow{2}{*}{Dataset} & \multicolumn{2}{c|}{Proposed}  & \multicolumn{2}{c|}{MemNet SR} & \multicolumn{2}{c|}{Laplacian SR} & \multicolumn{2}{c|}{SRCNN} \\ 
 
  & PSNR & SSIM & PSNR & SSIM & PSNR & SSIM & PSNR & SSIM\\
 \hline

Set-5 & 26.19 &	0.85 &	25.85 &	0.83 &	25.27 &	0.83 &	23.74 &	0.81\\
\hline
Set-14 & 24.21 &	0.79  &	24.1 &	0.77 &	24.15 &	0.77 &	23.04 &	0.76 \\
\hline
Urban-100   & 24.02 &	0.77  &	23.57 &	0.78  &	23.76 &	0.78 &	21.96	& 0.75 \\
\hline

\end{tabular}
\end{table*}

\begin{table*}[ht] 
\centering

 \caption{\label{tab:table_modifications}Model performance after ablation of certain components} 
 
 \begin{tabular}{|l | c| c| c| c|c| c| c| c| c|} 
 \hline
\multirow{2}{*}{Dataset} & \multicolumn{2}{c|}{\small{Baseline}}  & \multicolumn{2}{c|}{\small{Global residual learning removed}} & \multicolumn{2}{c|}{\small{Local Dense Connection removed}} & \multicolumn{2}{c|}{\small{Local residual learning removed}} \\ 
 
  & PSNR & SSIM & PSNR & SSIM & PSNR & SSIM & PSNR & SSIM\\
 \hline

Set 5 &  26.19 &	0.85 &	21.55 &	0.72  &	13.45 &	0.45 &	18.31 &	0.54 \\
\hline

Set 14 &  24.21 &	0.79  &	21.36 &	0.68 &	12.84 &	0.38 &	18.59 &	0.51\\
\hline
Urban-100  & 22.02 & 0.75  & 20.78 &	0.69  &	13.12 &	0.41 &	18.72 &	0.56\\
\hline

\end{tabular}
\end{table*}

The model was trained and tested with different degradation scales, such as 2x, 3x, and 4x. A comparison was also made with various other state-of-the-art models across different datasets. The efficiency of a few components in the architecture was also studied to understand their impact on the overall architecture.

Table \ref{tab:table_scale} shows the quantitative performance of the proposed model for different scales of LR images. It can be clearly seen that the performance of the model is better on 2x down-sampled images compared to others across all the datasets. This is due to the fact that the highly degraded images are less accurately converted to high-resolution images. A particular loss in details is bound to happen. The SSIM metric is around 0.90, while the PSNR is around 30 for 2x scaling, and the SSIM value is 0.85 for 4x scaling. The performance seemed to vary for different datasets, but a similar pattern was seen.

The performance of the proposed model with various state-of-the art models is shown in Table \ref{tab:table_model}. The comparison was made with SRCNN  \cite{srcnn}, MemNet \cite{tai2017memnet} and Laplacian SR \cite{lai2017deep}. SR CNN uses the pre-upsampling layer and a sparse encoding mechanism. MemNet uses the recursive learning mechanism mentioned in the literature survey along with the persistent memory mechanism. Laplacian SR uses a pyramid mechanism where at each level, feature maps are taken and high-frequency residual maps are predicted. Though other state-of-the-art networks exist, their performance was compared only with a selected few since they had a more distinct architecture than the proposed one and also due to the limitations of computation resources and time. The comparison was only performed with 4x scale degradation across all the models. The RDN that was used consists of only a lesser number of RDB blocks due to the training time limitation. However, it could be seen that the proposed model outperforms all the other state of the art models. From this, it is understandable that the dense residual learning-based architecture performs better than others.

Table \ref{tab:table_modifications} shows the performance of the model with some features removed from the architecture. A comparison was made with Baseline by removing the global residual learning functionality, the dense feature connection, and the local residual learning individually. Each of these modules or functionalities was removed individually, trained, and tested. From the results presented in the table, it could be inferred that the local dense connection was one of the important functionality since, without it, the model was not able to convert LR images to HR efficiently, which led to very poor SSIM and PSNR values. Local residual learning also seems to be an important functionality since removing it led to PSNR/SSIM values of around 18/0.54, while the baseline model was much better at around 26/0.85. This clearly shows the performance of the individual modules. Then global residual learning didn't lead to performance degradation as much as local dense connections and local residual learning. The global residual learning has a SSIM/PSNR value of 0.72/21.55, which is comparatively less than the baseline but not as important as the other two modules.

\section{Conclusion:}
This paper analyzes residual dense network architecture for single image super resolution, where novelty comes from residual dense blocks. A contiguous memory mechanism is achieved in RDB, where each block is connected to all the previous blocks in the module. Local dense feature fusion within each RDB helps with context and preserves local information. Local residual learning helps with the flow of context and feature knowledge within the local regions. All features are extracted in the LR space, leading to better performance with fewer computational resources. The proposed model uses both local and global features to effectively perform superresolution. A comparison was made with various state-of-the-art models. Also, the model was trained and tested for different degradation scales. The performance and importance of each module in the architecture were also evaluated.
 
{\small
\bibliographystyle{ieee}
\bibliography{egbib}
}

\end{document}